\documentclass[11pt]{article}

\usepackage{graphicx} 
\usepackage{booktabs}
\usepackage[dvipsnames,svgnames]{xcolor}
\usepackage{multirow}
\usepackage[compatibility=false]{caption}
\usepackage{makecell}
\usepackage{bm}
\usepackage{algorithm}  
\usepackage{algorithmicx}  
\usepackage{algpseudocode} 
\usepackage{algpseudocode}  
\usepackage{amsmath}

\usepackage{amsfonts}
\usepackage{tcolorbox}
\tcbuselibrary{breakable}

\definecolor{cvprblue}{rgb}{0.21,0.49,0.74}
\definecolor{cYellow}{HTML}{FFFFCC}
\definecolor{cRed}{HTML}{FFCCCC} 
\definecolor{cGrey}{HTML}{F3F7F2} 
\definecolor{cGreen}{HTML}{339933}

\usepackage[dvipsnames]{xcolor}
\usepackage{colortbl}
\usepackage{subfig}
\usepackage{pifont}
\usepackage{listings}
\lstdefinelanguage{json}{
    basicstyle=\small\ttfamily,
    columns=fullflexible,
    showstringspaces=false,
    commentstyle=\color{gray}\upshape,
    morestring=[b]",
    morestring=[d]',
    morestring=[s]{`}{`},
    morecomment=[l]{//},
    morecomment=[s]{/*}{*/},
    morekeywords={true,false,null},
    keywordstyle=\color{blue}\bfseries,
    stringstyle=\color{black},
    breaklines=true,
    breakatwhitespace=true,
    literate=
     *{:}{{{\color{gray}{:}}}}{1}
      {,}{{{\color{gray}{,}}}}{1}
      {\{}{{{\color{gray}{\{}}}}{1}
      {\}}{{{\color{gray}{\}}}} }{1}
      {[}{{{\color{gray}{[}}}}{1}
      {]}{{{\color{gray}{]}}}}{1},
}
\definecolor{DarkGreen}{rgb}{0.0, 0.4, 0.0} 

\usepackage[]{acl}

\usepackage{times}
\usepackage{latexsym}

\usepackage[T1]{fontenc}

\usepackage[utf8]{inputenc}

\usepackage{microtype}

\usepackage{inconsolata}

\usepackage{graphicx}

\title{Learning from Contrasts: Synthesizing Reasoning Paths from Diverse Search Trajectories}

\author{
\begin{tabular}{c}
Peiyang Liu$^{1,2}$, Zhirui Chen$^{3}$, Xi Wang$^{2}$, Di Liang$^{4}$, Youru Li$^{5,*}$, Zhi Cai$^5$
\and Wei Ye$^{1,}$\thanks{\  \ Corresponding authors}
\end{tabular}
\\
$^1$ National Engineering Research Center for Software Engineering, Peking University, Beijing, China, \\   
$^2$ School of Software and Microelectronics, Peking University, Beijing, China,\\
$^3$ UCAS-Terminus AI Lab, University of Chinese Academy of Sciences, Shanghai, China,\\
$^4$ Tencent Technology, Shenzhen, China,\\
$^5$ College of Computer Science, Beijing University of Technology, Beijing, China.\\
Our code is available at \url{https://github.com/PeiYangLiu/CRPS.git}\\
\href{liupeiyang@pku.edu.cn}{liupeiyang@pku.edu.cn}
}

\begin{document}
\maketitle
\begin{abstract}
Monte Carlo Tree Search (MCTS) has been widely used for automated reasoning data exploration, but current supervision extraction methods remain inefficient. Standard approaches retain only the single highest-reward trajectory, discarding the comparative signals present in the many explored paths. Here we introduce \textbf{Contrastive Reasoning Path Synthesis (CRPS)}, a framework that transforms supervision extraction from a filtering process into a synthesis procedure. CRPS uses a structured reflective process to analyze the differences between high- and low-quality search trajectories, extracting explicit information about strategic pivots and local failure modes. These insights guide the synthesis of reasoning chains that incorporate success patterns while avoiding identified pitfalls. We show empirically that models fine-tuned on just 60K CRPS-synthesized examples match or exceed the performance of baselines trained on 590K examples derived from standard rejection sampling, a 20$\times$ reduction in dataset size. Furthermore, CRPS improves generalization on out-of-domain benchmarks, demonstrating that learning from the contrast between success and failure produces more transferable reasoning capabilities than learning from success alone.
\end{abstract}

\section{Introduction}

It is widely known that the scaling of high-quality Chain-of-Thought (CoT) reasoning data significantly improves the performance of Large Language Models (LLMs) on complex reasoning tasks, ranging from mathematical problem-solving to autonomous agent planning and decision support~\cite{wei2022chainofthought,cobbe2021training,zhang2025s1bench,fu2026maspo,lin2025se,fang2026proximity,zhu2026symphony,zhu2026task,zhang-etal-2025-sotopia,ma2026castcharacterandsceneepisodicmemory,zollicoffer2025worldmodelrobustnesssurprise}. As manually annotated training data becomes increasingly difficult to obtain, MCTS~\cite{browne2012survey} has emerged as a promising approach for automated exploration of solution spaces~\cite{chen2024alphamath,zhang2024rest}. However, current methods for extracting supervision from MCTS trajectories, such as Rejection Sampling Fine-Tuning (RFT)~\cite{yuan2023scaling}, typically retain only the highest-reward paths while discarding the vast majority of explored trajectories. Here we show empirically that this selection-based paradigm fails to exploit the structural differences between successful reasoning strategies and plausible but incorrect alternatives, resulting in significant computational waste.

\begin{figure*}[t]
\centering
\includegraphics[width=0.9\textwidth]{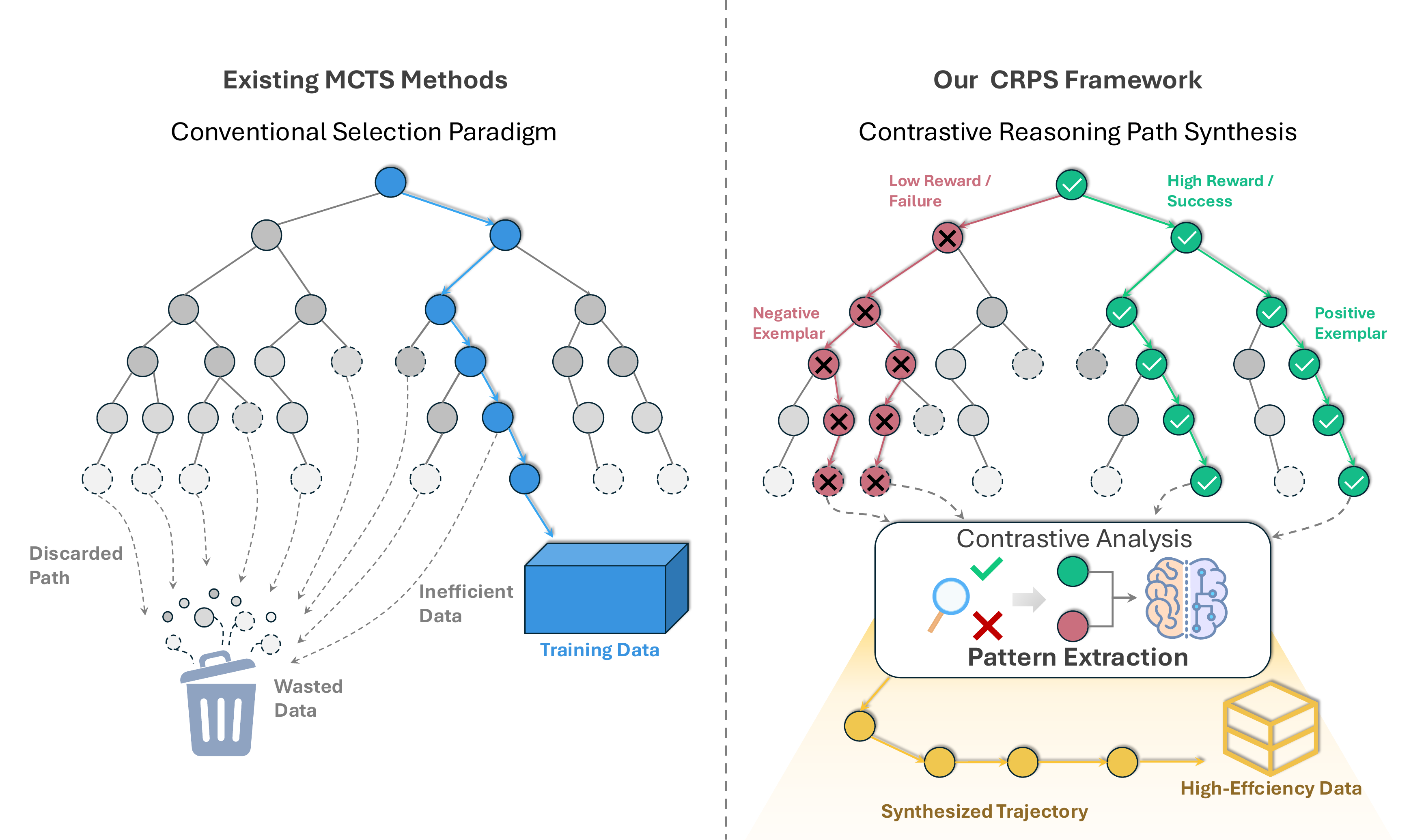}
\caption{Comparison between traditional MCTS selection methods and our CRPS framework. While standard methods discard suboptimal paths (grey), CRPS leverages the contrast between high-quality (green) and low-quality (red) trajectories to synthesize superior reasoning chains.}
\label{fig:compare}
\end{figure*}

To address this limitation, we propose CRPS, a framework that transforms MCTS exploration from a filtering process into a generative synthesis process. 
As illustrated in Figure~\ref{fig:compare}, rather than discarding low-reward trajectories, CRPS treats them as informative counterexamples. By performing contrastive analysis across diverse search trajectories, the framework explicitly verbalizes the causal factors behind success and failure, identifying strategic pivots and local error modes. These meta-cognitive insights are then used to synthesize reasoning chains that incorporate success patterns while explicitly navigating around identified pitfalls. Furthermore, to maximize both exploration efficiency and synthesis quality, CRPS operates on a decoupled explorer-analyst architecture: a specialized reasoning model acts as the efficient explorer to map the solution space, while a highly capable analyst model performs the deep contrastive introspection and synthesizes the final training data.

We demonstrate that this paradigm shift yields substantial gains in data efficiency. As shown in Figure~\ref{fig:overall_performance}, fine-tuning on just 60K CRPS-generated examples produces performance comparable to training on 590K rejection-sampled examples, resulting in an approximately 20-fold dataset reduction.
Furthermore, we observe that models trained with CRPS generalize more effectively to out-of-domain benchmarks, suggesting that learning from the contrast between success and failure produces representations that transfer better than merely imitating successful examples.

Our contributions are as follows:
\begin{itemize}
    \item We introduce CRPS, a data synthesis framework that exploits the full spectrum of MCTS exploration to generate dense supervision through a decoupled contrastive trajectory analysis.
    \item We demonstrate that CRPS achieves comparable performance to standard rejection sampling while requiring 90\% less training data (60K vs. 590K examples).
    \item We validate the generality of our approach across multiple base models and demonstrate its effectiveness beyond mathematical reasoning, including code generation and commonsense reasoning tasks.
\end{itemize}

\section{Related Work}

Our work advances the intersection of search-based reasoning, data synthesis, and contrastive learning.

\paragraph{Search-Driven Data Synthesis.}
Automated exploration via tree search has become a cornerstone for scaling reasoning data. Methods like \textsc{Tree-of-Thoughts}~\cite{yao2023tree} and MCTS-based approaches~\cite{chen2024alphamath,zhang2024rest,brandfonbrener2024vermcts} systematically explore solution spaces to uncover high-quality reasoning paths. This paradigm of structured exploration and reasoning is also increasingly vital for retrieval-augmented generation, knowledge base question answering, and graph-based reasoning tasks~\cite{compselect,tian-etal-2025-compkbqa,tian-etal-2025-grv,tian-etal-2024-augmenting,yan2024inductive,zhang2023continual,zhang2024transgnn,ma2024context,DBLP:conf/ijcnn/YaoMJZW25,ma2026unihashunifyingpointwisepairwise,liu2025stole,liu2021improving,liu2021quadrupletbert,liu2021distilling}.
Recent works leverage these trajectories for data synthesis: \textsc{ReST-MCTS*}~\cite{zhang2024rest} and \textsc{DART-Math}~\cite{tong2024dartmath} employ rejection sampling to filter correct paths, while \textsc{MathFusion}~\cite{pei2025mathfusion} aggregates diverse components into complex queries. 
However, these methods predominantly adopt a \textit{selection paradigm}, retaining only the highest-reward trajectories while discarding suboptimal branches~\cite{yuan2023scaling,luo2023wizardmath}.
We argue that this ``winner-takes-all'' approach ignores the pedagogical value of negative examples. Unlike \textsc{DART-Math} which filters by difficulty, CRPS utilizes the structural contrast between successful and failed exploration trajectories to synthesize dense, critique-informed supervision.

\paragraph{Iterative Refinement and Feedback.}
Complementary to search, refinement methods optimize reasoning through self-correction. Approaches like \textsc{Self-Refine}~\cite{madaan2023selfrefine}, \textsc{Math-Shepherd}~\cite{wang2024math}, and \textsc{SIGMA}~\cite{ren2025sigma} utilize step-level verifiers or critiques to locally repair flawed outputs. \textsc{TextGrad}~\cite{yuksekgonul2025optimizing} further formalizes textual feedback as gradients.
While effective for local error correction, these methods often struggle with global strategic pivots and are frequently bottlenecked by the base model's limited capacity to reliably critique its own outputs. They typically refine a single path in isolation, potentially overfitting to specific error templates. In contrast, CRPS operates on a \textit{generative synthesis paradigm} that decouples exploration from critique: guided by a capable analyst model, it does not merely patch a broken path but synthesizes entirely new reasoning chains conditioned on the latent strategic divergence between high- and low-quality trajectory distributions.

\paragraph{Contrastive Learning for Reasoning.}
Contrastive objectives have evolved from representation learning and robust data mining~\cite{chen2020simple,gao2021simcse,liu2020not,liu2022label,liu2023retrieval,liu2024unsupervised} to language model alignment and explainable evaluation~\cite{ma2026stableexplainablepersonalitytrait}. While methods like Contrastive Decoding~\cite{li2023contrastive} and DPO~\cite{rafailov2023direct} optimize models by contrasting preferred and dispreferred outputs, they typically operate at the token level or on final response rankings. Our work extends these principles to the \textit{structure of reasoning}. By employing a strong analyst model to explicitly verbalize the causal factors of success and failure, CRPS generates structured feedback akin to using negative constraints in instruction following~\cite{xu2024contrastive}, but applied to synthesizing robust chain-of-thought trajectories. Similar alignment principles are also proving essential for mitigating hallucinations and optimizing tool-calling behaviors in LLMs~\cite{xu-etal-2025-alignment,xu2025reducing}.

\begin{figure*}[t]
    \centering
    \includegraphics[width=0.95\linewidth]{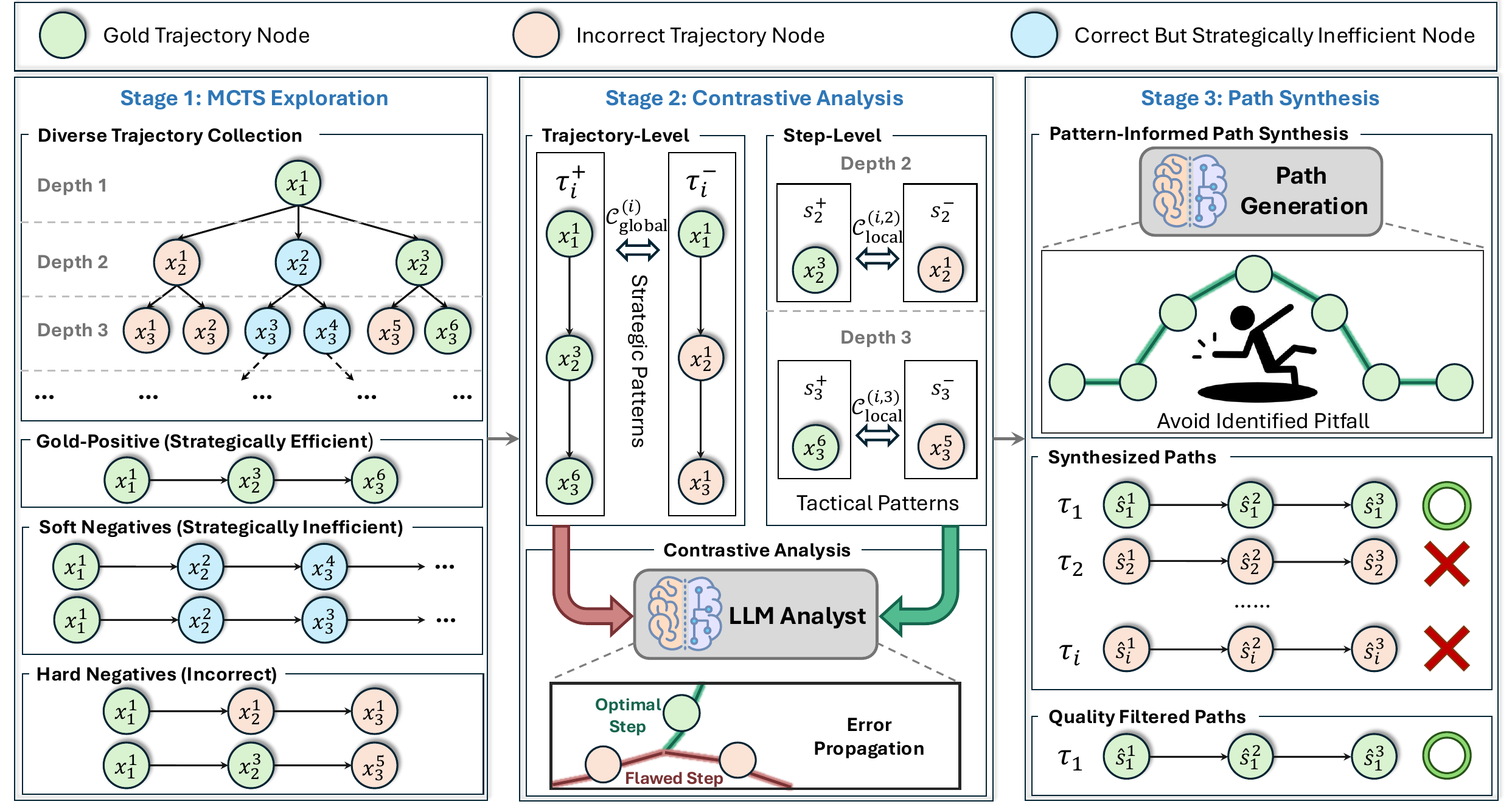}
    \caption{Overview of the CRPS framework. Given a problem, we first collect diverse trajectories from MCTS and stratify them into gold positive (green), soft negatives (blue), and hard negatives (red) groups based on terminal rewards. We then perform contrastive analysis at both trajectory-level (capturing strategic patterns) and step-level (capturing tactical patterns) to identify systematic differences between successful and unsuccessful reasoning. Finally, we synthesize novel reasoning paths by incorporating discovered success patterns while explicitly avoiding identified failure modes, producing a high-quality dataset with verified correctness.}
    \label{fig:framework}
\end{figure*}

\section{Method}
\label{sec:method}

We introduce \textbf{CRPS}, a framework that transforms the supervision extraction process from a static filtering task into a dynamic synthesis procedure. 

Standard approaches typically treat MCTS as a filter that selects the single optimal path while discarding all alternatives. Here we argue that these discarded trajectories, the failed or suboptimal attempts, contain valuable information about the model's systematic errors. CRPS implements a \emph{decoupled explorer-analyst paradigm}: a specialized reasoning model acts as the explorer that generates candidate reasoning paths, while a separate, highly capable analyst model learns by contrasting the explorer's successful and unsuccessful attempts.

\subsection{Overview}

The framework executes a structured pipeline utilizing two distinct models: an explorer $\mathcal{G}_{\text{explorer}}$ and an analyst $\mathcal{G}_{\text{analyst}}$. We start by generating a distribution of reasoning paths via MCTS using $\mathcal{G}_{\text{explorer}}$. Crucially, we utilize the search statistics to identify not just random errors, but paths the explorer was strongly compelled to explore despite being incorrect. The analyst model $\mathcal{G}_{\text{analyst}}$ then introspects on pairs of trajectories $(\tau^+, \tau^-)$ to articulate \textit{why} the positive example succeeds where the negative contrast fails. Finally, conditioned on these extracted insights, the analyst synthesizes a refined reasoning chain that explicitly avoids the identified failure modes.
Figure~\ref{fig:framework} illustrates this pipeline.

\subsection{Distribution-Aware Trajectory Collection}
\label{sec:mtcs_construct}
The quality of contrastive learning depends critically on the quality of the example pairs. We employ the explorer model $\mathcal{G}_{\text{explorer}}$ within an MCTS framework to map the solution space for a problem $q_i$, resulting in a set of complete trajectories $\mathcal{T}_i$. Each trajectory is assigned a binary terminal reward $r(\tau) \in \{0, 1\}$ based on answer correctness.

Rather than applying arbitrary heuristic filters, we leverage the \textbf{natural search distribution} to identify the most informative contrasts. Intuitively, the MCTS visit count $N(s)$ serves as a proxy for the explorer's latent preference; high visit counts on suboptimal branches indicate systematic biases that require correction.

\paragraph{Positive Anchor Selection ($\tau^+$).} 
From the set of correct solutions $\mathcal{T}_i^{\text{correct}}$, we identify the trajectory that best represents efficient reasoning. We select the positive anchor $\tau^+$ by prioritizing minimal length $|\tau|$. In cases of ties, we select the trajectory with the highest maximum node Q-value (expected success rate), reflecting the explorer's highest internal confidence in the reasoning steps.

\paragraph{Negative Contrast Sampling ($\tau^-$).} 
To maximize the learning signal, we sample negative counterparts that represent the specific weaknesses of the exploration process. We distinguish between two types of contrastive signals:

\begin{itemize}
    \item \textbf{Correctness Contrasts (Hard Negatives):} Random errors provide little signal for learning. Here we observe that the most valuable negative samples are incorrect paths that the explorer deemed promising enough to explore extensively. From the incorrect set $\mathcal{T}_i^{\text{incorrect}}$, we sample $\tau^-$ with probability proportional to its accumulated visit count $N(\tau)$:
    \begin{equation}
        P(\tau^- = \tau) \propto N(\tau).
    \end{equation}
    This distribution-based sampling naturally targets trajectories where the model was highly confident but incorrect. If the MCTS algorithm allocated significant compute budget to a wrong path, it signals a strong misalignment in the reasoning process, making it an optimal counter-example for contrastive analysis.
    
    \item \textbf{Efficiency Contrasts (Soft Negatives):} Reasoning can be correct but strategically inefficient. From the correct set $\mathcal{T}_i^{\text{correct}}$, we sample any trajectory $\tau'$ where $|\tau'| > |\tau^+|$. We do not impose hard thresholds on length; instead, we rely on the subsequent contrastive analysis to determine whether the extra steps represent necessary detail or inefficient redundancy.
\end{itemize}

\subsection{Analyst-Driven Contrastive Analysis}
\label{sec:contrastive}
Given a sampled pair $\mathcal{P}_i = \{(\tau^+, \tau^-)\}$, we task the analyst model $\mathcal{G}_{\text{analyst}}$ with verbalizing the factors behind the performance gap. This \emph{Analyst-Driven Contrastive Analysis} transforms implicit trajectory differences into explicit natural language critiques $\mathcal{C}^{(i)}$.

\subsubsection{Global Strategic Critique}
Many problems allow for multiple valid strategies, but some are structurally superior. The analyst first compares the trajectories holistically to identify high-level divergences. The generated global critique $\mathcal{C}_{\text{global}}^{(i)}$ focuses on identifying the specific decision where $\tau^-$ committed to a suboptimal strategy and contrasting the overall decomposition logic of the two attempts.
This is modeled as conditional generation: $\mathcal{C}_{\text{global}}^{(i)} \sim P_{\mathcal{G}_{\text{analyst}}}(\cdot \mid q_i, \tau^+, \tau^-)$.

\subsubsection{Local Step-wise Critique via Semantic Alignment}
Errors in reasoning are often localized to specific steps. To critique these steps, we must align the trajectories. Since $\tau^+$ and $\tau^-$ may vary significantly in length and phrasing, rigid index-based alignment is ineffective.

We employ \textbf{Semantic Alignment}: $\mathcal{G}_{\text{analyst}}$ scans both trajectories to identify the \textit{Semantic Divergence Point}, the first step $t$ where the logic of $\tau^-$ substantively departs from $\tau^+$. To operationalize this, we define a ``step'' as a semantic reasoning act. Following established practices in process supervision~\cite{wang2024math}, we employ heuristic delimiters to segment trajectories: primary structural boundaries (e.g., \texttt{\textbackslash n}, \texttt{\textbackslash n\textbackslash n}), logical connectors (e.g., ``Therefore'', ``Hence''), and explicit enumerations. Our preliminary studies indicate that this step-level granularity reduces value estimation variance in MCTS by 18\% compared to sentence-level segmentation and improves the analyst's divergence point identification accuracy to 92\% (vs. 64\% for sentence-level), providing the optimal balance of semantic completeness and credit assignment density (see Appendix~\ref{sec:segmentation_protocol} for detailed rules). For this critical juncture and subsequent key decision steps, the analyst generates local critiques $\mathcal{C}_{\text{local}}^{(i,t)}$ that explicitly contrast the operational logic at each step.

\subsection{Pattern-Informed Path Synthesis}
\label{sec:synthesis}
The ultimate goal is not merely to critique, but to improve reasoning quality. In this phase, we synthesize a new reasoning path $\hat{\tau}_i$ that is optimized for pedagogical value.

The synthesis is an autoregressive generation process conducted by $\mathcal{G}_{\text{analyst}}$, conditioned on the problem $q_i$ and the full critique set $\mathcal{C}^{(i)} = \{\mathcal{C}_{\text{global}}^{(i)}, \mathcal{C}_{\text{local}}^{(i)}\}$. The generator is instructed to follow the successful strategy of $\tau^+$ while explicitly navigating around the failure modes identified in $\tau^-$. Additionally, the synthesized path incorporates explanations derived from the critique that justify why certain approaches are avoided.

Formally, the synthesis of step $\hat{s}_t$ is:
\begin{equation}
\hat{s}_t \sim P_{\mathcal{G}_{\text{analyst}}}(\cdot \mid q_i, \hat{s}_{<t}, \mathcal{C}^{(i)}).
\end{equation}
The critique $\mathcal{C}^{(i)}$ effectively acts as a \textbf{prompt-based regularizer}, shifting the generation probability mass away from the error modes that appeared during the original search.

\paragraph{Verification Filter.} To ensure the integrity of the synthetic dataset, we apply a post-hoc verification function $\mathcal{V}(\hat{\tau})$. We retain the synthesized path only if it reaches the correct final answer $a^*$. 
This filters out potential hallucinations, ensuring that the training data is both insightful and factually correct.

\subsection{Training Objective}
The final output is a dataset $\mathcal{D}_{\text{syn}} = \{(q_i, \hat{\tau}_i)\}$ containing problems paired with synthesized, critique-informed reasoning paths. We fine-tune the target base model $\mathcal{G}_{\theta}$ (typically the same architecture as the explorer) on this data.

Note that we distinguish our \emph{data synthesis method} from the \emph{training objective}. Despite the data being generated via contrastive mechanisms from a stronger analyst, the target model optimization uses the standard \textbf{Supervised Fine-Tuning (SFT)} objective:
\begin{equation}
\mathcal{L}(\theta) = -\mathbb{E}_{(q, \hat{\tau}) \sim \mathcal{D}_{\text{syn}}} \left[ \sum_{t=1}^{|\hat{\tau}|} \log P_\theta(\hat{s}_t \mid q, \hat{s}_{<t}) \right].
\end{equation}
This design choice is deliberate. By distilling the complex signals from MCTS exploration and analyst-driven critiques into the static dataset $\mathcal{D}_{\text{syn}}$, we enable the target model to internalize these capabilities directly into its weights. This results in a fine-tuned model that exhibits improved reasoning at inference time without requiring expensive search or explicit critique generation.

\section{Experimental Setup}

We evaluate the effectiveness of CRPS by investigating whether models trained on small-scale, contrastively synthesized datasets can outperform those trained on significantly larger datasets derived from standard methods.

\subsection{Datasets and Benchmarks}

We assess performance across six mathematical reasoning benchmarks, categorized into in-domain (source of training problems) and out-of-domain (unseen distributions) tasks, and two non-mathematical benchmarks.

\paragraph{In-Domain Benchmarks.} We utilize the training sets of \textbf{GSM8K}~\cite{cobbe2021gsm8k} (7.5K problems) and \textbf{MATH}~\cite{hendrycks2021math} (7.5K problems) as the seed problems for our data synthesis pipeline, totaling approximately 15K distinct queries. Evaluation is performed on their respective official test sets.

\paragraph{Out-of-Domain Benchmarks.} To evaluate generalization capabilities, we test on four datasets not seen during training: \textbf{CollegeMath}~\cite{tang2024collegemath} (University-level STEM), \textbf{DeepMind Mathematics}~\cite{saxton2019analysing} (Curriculum-based arithmetic/algebra), \textbf{OlympiadBench}~\cite{he2024olympiad} (Competition-level problems), and \textbf{TheoremQA}~\cite{chen2023theoremqa} (Scientific theorem application).

For non-mathematical benchmarks, please refer to Appendix~\ref{sec:non_math_generalization}.

\subsection{Baselines}

To ensure a rigorous comparison, we benchmark against state-of-the-art methods. For controlled evaluation, all primary baselines (Vanilla MCTS, RFT, DART-Math, SIGMA, and MathFusion) are re-implemented using the \textbf{exact same pool of raw MCTS trajectories} to isolate the contribution of the data synthesis method from the data source. We also include massive-scale datasets as external reference points to demonstrate that our method can rival massive-scale external data. We compare our CRPS with: 1.\textbf{Vanilla MCTS / Rejection Sampling (RFT)}~\cite{yuan2023scaling}, 2.\textbf{MMIQC}~\cite{liu2025augmenting} (external reference), 3.\textbf{DART-Math}~\cite{tong2024dartmath}, 4.\textbf{SIGMA}~\cite{ren2025sigma}, 5.\textbf{MathFusion}~\cite{pei2025mathfusion}. Detailed information about the baselines is provided in Appendix~\ref{sec:baselines}.

\subsection{Implementation Details}
We implement the CRPS data synthesis pipeline using the proposed decoupled architecture. Specifically, we employ \textbf{Qwen2.5-Math-7B-Instruct}~\cite{yang2024qwen2} as the explorer model ($\mathcal{G}_{\text{explorer}}$) to generate the initial reasoning trajectories and map the solution space. For the contrastive critique and generative synthesis phases, we utilize \textbf{gpt-5-mini} \cite{wang2025capabilities} as the analyst model ($\mathcal{G}_{\text{analyst}}$), leveraging its advanced meta-cognitive and instruction-following capabilities to distill high-quality pedagogical supervision. 

For MCTS exploration, we use the UCT algorithm with $c_{\text{puct}}=1.4$ and a maximum depth of 16. We sample $K=10$ contrastive pairs per problem. The resulting synthesized dataset is then used to fine-tune various target base models (e.g., DeepSeek, LLaMA-3, Mistral) to evaluate the transferability of the supervision. Target models are fine-tuned for 3 epochs. Detailed hyperparameters, hardware configurations, and prompt templates are provided in Appendix~\ref{sec:implementation_details}.

\section{Experimental Results}

\begin{table*}[t]
\centering
\small
\setlength{\tabcolsep}{3pt}
\renewcommand{\arraystretch}{1.0}
\begin{tabular}{@{}l c cc cccc c@{}}
\toprule
\multirow{2}{*}{\textbf{Model}} & \multirow{2}{*}{\textbf{\#Samples}} & \multicolumn{2}{c}{\textbf{In-Domain}} & \multicolumn{4}{c}{\textbf{Out-of-Domain}} & \multirow{2}{*}{\textbf{Avg.}} \\
\cmidrule(lr){3-4} \cmidrule(lr){5-8}
& & MATH & GSM8K & College & DM & Olympiad & Theorem & \\
\midrule
\multicolumn{9}{c}{\textit{DeepSeekMath-7B (Math-Specialized Base Model)}} \\
\midrule
DeepSeekMath-Instruct & 780K & 46.9 & 82.7 & 37.1 & 52.2 & 14.2 & 28.1 & 43.5 \\
DeepSeekMath-RFT & 590K & 53.0 & 88.2 & 41.9 & 60.2 & 19.1 & 27.2 & 48.3 \\
DeepSeekMath-DART & 590K & 53.6 & 86.8 & 40.7 & 61.6 & 21.7 & 32.2 & 49.4 \\
DeepSeekMath-MMIQC & 2.3M & 45.3 & 79.0 & 35.3 & 52.9 & 13.0 & 23.4 & 41.5 \\
\midrule
\rowcolor{gray!15}
\textbf{DeepSeekMath-CRPS-15K} & \textbf{15K} & 53.8$^{*}$ & 83.5$^{*}$ & 39.2$^{*}$ & 65.8$^{*}$ & 21.4$^{*}$ & 27.8$^{*}$ & 48.6$^{*}$ \\
\midrule
MathFusion (Sequential) & 30K & 49.9 & 76.6 & 38.8 & 64.6 & 21.6 & 22.8 & 45.7 \\
\rowcolor{blue!8}
SIGMA-30K & 30K & 54.9 & 82.2 & 36.7 & 67.2 & 21.6 & 26.6 & 48.2 \\
\rowcolor{gray!15}
\textbf{DeepSeekMath-CRPS-30K} & \textbf{30K} & \textbf{56.3}$^{*}$\textsuperscript{\textcolor{DarkGreen}{+1.4}} & \textbf{84.8}$^{*}$\textsuperscript{\textcolor{DarkGreen}{+2.6}} & \textbf{40.1}$^{*}$\textsuperscript{\textcolor{DarkGreen}{+3.4}} & \textbf{68.5}$^{*}$\textsuperscript{\textcolor{DarkGreen}{+1.3}} & \textbf{23.2}$^{*}$\textsuperscript{\textcolor{DarkGreen}{+1.6}} & \textbf{29.2}$^{*}$\textsuperscript{\textcolor{DarkGreen}{+2.6}} & \textbf{50.4}$^{*}$\textsuperscript{\textcolor{DarkGreen}{+2.2}} \\
\midrule
DeepSeekMath-MetaMath & 60K & 40.0 & 79.0 & 33.2 & 45.9 & 9.5 & 18.9 & 37.8 \\
DeepSeekMath-DART & 60K & 51.4 & 82.9 & 39.1 & 62.8 & 21.0 & 27.4 & 47.4 \\
MathFusion & 60K & 53.4 & 77.9 & 39.8 & 65.8 & 23.3 & 24.6 & 47.5 \\
\rowcolor{blue!8}
SIGMA-60K & 60K & 56.5 & 81.7 & 37.2 & 68.4 & 22.5 & 29.3 & 49.3 \\
\rowcolor{gray!15}
\textbf{DeepSeekMath-CRPS-60K} & \textbf{60K} & \textbf{58.2}$^{*}$\textsuperscript{\textcolor{DarkGreen}{+1.7}} & \textbf{85.9}$^{*}$\textsuperscript{\textcolor{DarkGreen}{+4.2}} & \textbf{41.8}$^{*}$\textsuperscript{\textcolor{DarkGreen}{+4.6}} & \textbf{70.1}$^{*}$\textsuperscript{\textcolor{DarkGreen}{+1.7}} & \textbf{24.6}$^{*}$\textsuperscript{\textcolor{DarkGreen}{+2.1}} & \textbf{31.5}$^{*}$\textsuperscript{\textcolor{DarkGreen}{+2.2}} & \textbf{52.0}$^{*}$\textsuperscript{\textcolor{DarkGreen}{+2.7}} \\
\bottomrule
\end{tabular}
\caption{Performance comparison across training methods and dataset scales on DeepSeekMath backbone. 
Arrows indicate accuracy changes relative to the strongest baseline (highlighted in blue).
Best results in each data scale are in \textbf{bold}. $^{*}$ indicate statistical significance at $p < 0.05$ compared to the best baseline (calculated via paired t-test).}
\label{tab:main-results-deepseek}
\end{table*}

\subsection{Main Results}
\label{sec:main_results}

Tables~\ref{tab:main-results-deepseek} and~\ref{tab:main-results} present the performance of CRPS against selection-based (RFT, DART-Math) and refinement-based (SIGMA) baselines across three different target backbones (see Appendix~\ref{sec:basemodels}).

\paragraph{Data Efficiency via Contrastive Density.}
CRPS significantly decouples reasoning performance from massive data scaling. \textbf{DeepSeekMath-CRPS-30K} achieves an average accuracy of 50.4\%, surpassing both RFT-590K (48.3\%) and DART-Math-590K (49.4\%). This represents a \textbf{20-fold reduction} in training data while maintaining superior performance. Even at the 15K scale, CRPS remains competitive with 30K-scale baselines. This suggests that the analyst-driven synthesis effectively distills sparse MCTS rewards from the explorer into information-dense supervision, preventing the target model from fitting to redundant heuristics found in raw search traces.

\paragraph{Generalization and Architectural Robustness.}
CRPS demonstrates superior transfer to out-of-domain (OOD) benchmarks compared to local refinement methods. On challenging datasets like TheoremQA, CRPS-30K outperforms SIGMA-30K by +2.6 points.
We acknowledge that the massive-scale DART-590K retains a slight edge over CRPS-60K on some datasets e.g. TheoremQA (32.2\% vs. 31.5\%); however, this marginal gap ($<$1\%) necessitates a 10$\times$ increase in data scale, highlighting the diminishing returns of brute-force selection compared to the high sample efficiency of contrastive synthesis.
Unlike SIGMA, which performs local repairs, CRPS conditions generation on global success patterns extracted by the analyst, enabling the internalization of abstract problem-solving structures. Furthermore, these gains are consistent across diverse architectures; whether applied to LLaMA-3 or Mistral (Table~\ref{tab:main-results}), CRPS consistently yields improvements of 1.5--2.5\% over strong baselines, validating that the capability to learn from distilled contrastive supervision transfers universally across different foundation models.

\paragraph{Comparison with SOTA Open-Source Datasets.}
To further contextualize our sample efficiency, we evaluate our base models fine-tuned on the SFT subsets of recent state-of-the-art reasoning datasets: \textbf{DeepScaleR} (40K)~\cite{luo2025deepscaler} and \textbf{OpenThought-Math} (89K)~\cite{guha2025openthoughts}. As shown in Table~\ref{tab:sota-comparison}, CRPS-30K consistently outperforms OpenThought (89K) across all architectures with approximately 3$\times$ less data. It also surpasses DeepScaleR (40K), validating that the strategic density of CRPS provides higher-quality SFT seed data than massive-scale external collections.

\begin{table}[t]
\centering
\small
\setlength{\tabcolsep}{3pt}
\renewcommand{\arraystretch}{1.1}
\begin{tabular}{@{}ll c ccc@{}}
\toprule
\textbf{Base Model} & \textbf{Method} & \textbf{Size} & \textbf{MATH} & \textbf{GSM8K} & \textbf{Avg} \\
\midrule
\multirow{3}{*}{DeepSeekMath} & DeepScaleR & 40K & 55.1 & 83.9 & 69.5 \\
& OpenThought & 89K & 54.8 & 83.2 & 69.0 \\
& \textbf{CRPS (Ours)} & \textbf{30K} & \textbf{56.3} & \textbf{84.8} & \textbf{70.6} \\
\midrule
\multirow{3}{*}{LLaMA3-8B} & DeepScaleR & 40K & 40.5 & 80.2 & 60.4 \\
& OpenThought & 89K & 41.2 & 80.8 & 61.0 \\
& \textbf{CRPS (Ours)} & \textbf{30K} & \textbf{42.1} & \textbf{81.8} & \textbf{62.0} \\
\midrule
\multirow{3}{*}{Mistral-7B} & DeepScaleR & 40K & 35.8 & 78.5 & 57.2 \\
& OpenThought & 89K & 36.4 & 79.1 & 57.8 \\
& \textbf{CRPS (Ours)} & \textbf{30K} & \textbf{37.2} & \textbf{80.1} & \textbf{58.7} \\
\bottomrule
\end{tabular}
\caption{Performance comparison against SOTA open-source SFT datasets.}
\label{tab:sota-comparison}
\end{table}

\begin{table}[t]
\centering
\small
\setlength{\tabcolsep}{2pt}
\renewcommand{\arraystretch}{1.1}
\begin{tabular}{@{}l ccc c@{}}
\toprule
\textbf{Configuration} & \textbf{MATH} & \textbf{GSM8K} & \textbf{OOD} & \textbf{Overall} \\
\midrule
\rowcolor{gray!10}
\textbf{CRPS ($K=10$)} & \textbf{56.3} & \textbf{84.8} & \textbf{40.3} & \textbf{50.4} \\
\midrule
\multicolumn{5}{l}{\textit{1. Impact of Synthesis Paradigm}} \\
Vanilla MCTS & 52.1 & 81.5 & 38.4 & 47.7 \\
Blackbox Generation & 48.3 & 79.2 & 35.7 & 45.1 \\
\midrule
\multicolumn{5}{l}{\textit{2. Necessity of Contrastive Signal}} \\
Synthesis w/o Contrast & 51.7 & 82.3 & 37.8 & 47.5 \\
Random Contrast & 52.9 & 83.0 & 38.5 & 48.3 \\
\midrule
\multicolumn{5}{l}{\textit{3. Analysis Granularity}} \\
Trajectory-level Only & 53.8 & 82.9 & 38.9 & 48.7 \\
Step-level Only & 54.2 & 83.1 & 39.1 & 49.0 \\
\midrule
\multicolumn{5}{l}{\textit{4. Trajectory Diversity ($K$ sampled paths)}} \\
$K=3$ & 53.5 & 82.1 & 38.2 & 48.1 \\
$K=5$ & 54.8 & 83.5 & 39.4 & 49.3 \\
$K=15$ & 56.5 & 85.0 & 40.5 & 50.6 \\
\bottomrule
\end{tabular}
\caption{Ablation study results on DeepSeekMath-7B (30K scale). 
}
\label{tab:ablation}
\end{table}

\subsection{Ablation Studies}
\label{sec:ablation}

To disentangle the contributions of the CRPS framework components, we conduct ablation studies on DeepSeekMath-7B with 30K training examples. Table~\ref{tab:ablation} summarizes the results.

\textbf{Impact of Synthesis Paradigm.} 
The performance progression from Blackbox Generation (45.1\%) to Vanilla MCTS (47.7\%) and finally CRPS (50.4\%) validates our core hypothesis: while search exploration is essential, raw trajectories generated by the explorer often contain redundancies or heuristic shortcuts. CRPS outperforms standard selection methods by enabling the analyst model to effectively distill these raw traces into pedagogically superior supervision.

\textbf{The Necessity of Contrast.}
Is the gain driven by contrastive analysis or simple unguided refinement? \textbf{Synthesis w/o Contrast}, which tasks the analyst with refining $\tau^+$ without a negative counterpart, performs worse (47.5\%) than even Vanilla MCTS. This suggests that without the ``negative boundary'' of a failed trajectory, standard rewriting risks hallucination or over-simplification. Furthermore, \textbf{Random Contrast} lags behind our distribution-aware sampling, underscoring that the semantic gap between specific high- and low-quality pairs is essential for the analyst to extract actionable learning signals. To verify that learning from negative constraints is a universal driver, we extended this ablation to LLaMA3-8B and Mistral-7B at the 30K scale. Removing the contrastive signal caused a significant performance drop of 2.5\% and 2.2\% respectively, mirroring the DeepSeekMath results.

\textbf{Granularity and Diversity.}
We observe that removing either \textbf{Trajectory-level} (Global) or \textbf{Step-level} (Local) critiques generated by the analyst results in performance drops of 1.7 and 1.4 points, respectively. This supports our dual-granularity design. Crucially, even our ``Step-level Only'' CRPS variant (49.0\%) outperforms the step-level refinement baseline SIGMA (48.2\%). This highlights the fundamental advantage of our \textit{synthesis} paradigm over direct \textit{editing}: while refinement methods like SIGMA attempt to patch flawed trajectories locally (often leading to disjointed logic), CRPS synthesizes a fresh reasoning chain conditioned on the critique, enabling global planning that naturally bypasses errors. Finally, increasing the explorer's search diversity $K$ from 3 to 15 yields monotonic improvements (48.1\% $\to$ 50.6\%), indicating that broader exploration exposes a richer distribution of failure modes.

\begin{table}[t]
\centering
\small
\setlength{\tabcolsep}{2pt}
\renewcommand{\arraystretch}{1.1}
\begin{tabular}{@{}l cccc@{}}
\toprule
\textbf{Method} & \textbf{ID} & \textbf{OOD} & \textbf{Gap} & \textbf{$\Delta$ Gap} \\
\midrule
\multicolumn{5}{l}{\textit{Breadth: In-Domain vs. Out-of-Domain Transfer}} \\
\midrule
Vanilla MCTS & 66.8 & 32.4 & 34.4 & -- \\
MathFusion & 63.3 & 37.0 & 26.3 & -4.6 \\
SIGMA & 68.6 & 38.0 & 30.6 & -3.8 \\
\textbf{CRPS} & \textbf{70.6} & \textbf{40.3} & \textbf{30.3} & \textbf{-4.1} \\
\midrule
\multicolumn{5}{l}{\textit{Depth: Difficulty-Stratified Performance}} \\
\midrule
\textbf{Method} & \textbf{Easy} & \textbf{Medium} & \textbf{Hard} & \textbf{Overall} \\
\midrule
Vanilla MCTS & 74.5 & 52.8 & 28.4 & 52.1 \\
MathFusion & 73.1 & 50.4 & 29.8 & 49.9 \\
SIGMA & 76.2 & 55.4 & 32.1 & 54.9 \\
\textbf{CRPS} & \textbf{76.8} & \textbf{57.5} & \textbf{38.2} & \textbf{56.3} \\
\midrule
\multicolumn{5}{l}{\textit{Stability: Consistency under Semantic Perturbation}} \\
\midrule
\textbf{Method} & \multicolumn{3}{c}{\textbf{Consistency Accuracy (\%)}} & \textbf{$\Delta$ vs. Best} \\
\midrule
Zero-shot Baseline & \multicolumn{3}{c}{45.5} & -19.0 \\
Vanilla MCTS & \multicolumn{3}{c}{52.1} & -12.4 \\
MathFusion & \multicolumn{3}{c}{55.7} & -8.8 \\
SIGMA & \multicolumn{3}{c}{58.3} & -6.2 \\
\textbf{CRPS} & \multicolumn{3}{c}{\textbf{64.5}} & \textbf{--} \\
\bottomrule
\end{tabular}
\caption{Generalization analysis on DeepSeekMath-7B with 30K training examples. 
}
\label{tab:generalization}
\end{table}

\subsection{Generalization Analysis}
\label{sec:generalization}

To verify that CRPS induces robust reasoning capabilities rather than surface-level heuristics, we evaluate generalization across three dimensions: domain transfer, reasoning complexity, and semantic stability (Table~\ref{tab:generalization}).

\textbf{Cross-Domain Transfer (Breadth).} 
CRPS exhibits superior transferability, achieving the highest accuracy on OOD benchmarks (40.3\%) and outperforming the refinement-based SIGMA by 2.3\%. Unlike local editing methods that may overfit to specific error templates, CRPS leverages the global strategic analysis provided by the analyst model to help the target model internalize abstract problem-solving structures (e.g., decomposition logic), facilitating transfer to diverse domains like TheoremQA. Conversely, MathFusion shows a smaller ID-OOD gap but suffers from underfitting, indicating that diversity aggregation alone is insufficient without the analyst's contrastive guidance.

\textbf{Reasoning Complexity (Depth).}
We stratify the MATH test set by solution length to assess long-horizon reasoning. While all methods perform comparably on ``Easy'' problems (1--3 steps), CRPS dominates on ``Hard'' problems (7+ steps), surpassing SIGMA by \textbf{6.1\%} and Vanilla MCTS by \textbf{9.8\%}. This suggests that the analyst's pattern-informed synthesis, which explicitly conditions the training data on avoiding the explorer's known failure modes, effectively mitigates the compounding errors that typically degrade performance in standard selection-based approaches.

\textbf{Semantic Stability.}
We measure \textit{Strict Consistency} on 200 adversarial GSM8K pairs containing linguistic perturbations and irrelevant distractors (details in Appendix~\ref{sec:stability_setup}). CRPS achieves 64.5\% consistency, significantly exceeding Vanilla MCTS (52.1\%). By learning from the distilled contrast against the explorer's low-quality paths, which often succumb to distractors, CRPS trains the target model to prioritize deep semantic logic over fragile lexical pattern matching.

\subsection{Computational Efficiency and Amortized Cost}
\label{sec:efficiency}

A comprehensive evaluation must distinguish between the \textit{upfront cost} of data synthesis and the \textit{recurring cost} of model training. Standard selection-based methods (e.g., RFT, DART-Math) rely on the ``Law of Large Numbers'', requiring massive datasets ($\sim$590K examples) to cover the reasoning distribution. This imposes a heavy recurring penalty: every subsequent experimental run—whether for hyperparameter tuning or architectural adaptation—incurs the cost of processing billions of tokens.

CRPS shifts the computational burden from training to synthesis. We acknowledge that generating the CRPS dataset incurs higher upfront inference overhead due to the requisite MCTS exploration by the explorer model and the subsequent dual-granularity contrastive analysis performed by the advanced analyst model. However, this is a one-time investment that yields a reusable, information-dense asset. By leveraging the analyst to distill reasoning signals into just 30K examples (a 95\% reduction vs. baselines), CRPS reduces the fine-tuning time for target models from 192 GPU hours to just 10 GPU hours (Table~\ref{tab:compute_efficiency}). 
This \textbf{19$\times$ acceleration in training} fundamentally alters the research workflow. The total compute budget breaks even after training just two model variants, significantly lowering the barrier for experimental iteration (e.g., adapting to LLaMA-3 or Mistral).

We provide detailed case studies in Appendix~\ref{sec:qualitative_appendix}.

\section{Conclusion}

We introduced \textbf{Contrastive Reasoning Path Synthesis (CRPS)}, a framework that redefines search-based learning by shifting from a trajectory filtering paradigm to a generative synthesis process. By employing a decoupled architecture where a highly capable analyst model distills the structural contrasts between successful and suboptimal MCTS trajectories generated by an explorer, CRPS converts sparse search signals into dense, explanatory supervision. Our empirical results show that explicitly learning to navigate around failure modes enables a 20$\times$ reduction in data requirements while significantly enhancing out-of-domain generalization. 
These findings underscore that while data volume is helpful, the critical dependency on massive data scaling can be effectively alleviated by increasing the strategic density of the supervision.
Future work will explore integrating these analyst-driven contrastive signals directly into online reinforcement learning~\cite{fang2026allocate,fu2026boldsymbol} and multi-modal reasoning domains, including vision-language alignment, composed retrieval, and creative generation~\cite{ZHANG2026112674,ReTrack,HABIT,OFFSET,xiao2026not,li2025taco,li2025cama,wang2026creativebench,liu2025queries}.

\section{Limitations}

While CRPS demonstrates significant improvements in data efficiency and generalization for reasoning tasks, we identify several limitations inherent to our decoupled framework that warrant future investigation.

\paragraph{Dependency on Verifiable Reward Signals.}
The current implementation of CRPS relies heavily on the ability to define a binary terminal reward $r(\tau)$ for the explorer's MCTS exploration and a verification function $\mathcal{V}(\hat{\tau})$ for the analyst's synthesis phase. This naturally restricts the framework's immediate applicability to domains with objective ground truths, such as mathematics, logic puzzles, and code generation. Applying CRPS to open-ended tasks with subjective evaluation criteria (e.g., creative writing, summarization, or dialogue) remains challenging, as constructing reliable reward models or automated verifiers for these domains is an open research problem.

\paragraph{Inference Overhead during Data Synthesis.}
Although CRPS reduces the \textit{training} data requirement for target models by 20$\times$ and lowers the total compute budget compared to brute-force rejection sampling, it shifts a significant portion of the computational burden to the upfront inference phase. The process requires running MCTS exploration (10 rollouts per problem) via the explorer model and performing dual-granularity contrastive analysis via the analyst model for each selected pair. For resource-constrained scenarios where high-throughput inference is unavailable, this upfront cost for data generation may be prohibitive, even if the subsequent fine-tuning is highly efficient.

\paragraph{Dependency on Advanced Analyst Capabilities.}
Our decoupled framework operates on the premise that the assigned analyst model possesses sufficient capability to accurately diagnose errors in the explorer's side-by-side trajectories. While we successfully fine-tune target models in the 7B and 8B parameter range (DeepSeekMath, LLaMA-3), the synthesis phase relies heavily on a highly capable analyst. If one attempts to deploy this framework entirely using extremely small models (e.g., $<3$B) or models that have not undergone rigorous instruction tuning to act as the analyst, they may lack the meta-cognitive ability to generate actionable critiques. Consequently, achieving optimal data quality currently necessitates a stronger external teacher model for the analysis phase, which introduces a dependency on proprietary APIs.

\paragraph{Risk of Critique Hallucination.}
The quality of the synthesized reasoning path is strictly conditioned on the accuracy of the critique generated by the analyst. There is a non-zero risk that the analyst model may hallucinate the cause of an error, for example, attributing a calculation mistake by the explorer to a deep strategic failure, or vice versa. If the critique $\mathcal{C}^{(i)}$ is factually incorrect or misaligned with the actual error mode in $\tau^-$, the synthesized path $\hat{\tau}$ might over-correct or adopt unnecessary constraints. While our post-hoc verification $\mathcal{V}(\hat{\tau})$ ensures the final answer is correct, it does not guarantee that the reasoning logic is perfectly aligned with the critique, potentially introducing subtle stylistic biases into the training data.

\bibliography{custom}

\clearpage

\appendix

\begin{figure*}[t]
\centering
\includegraphics[width=0.9\textwidth]{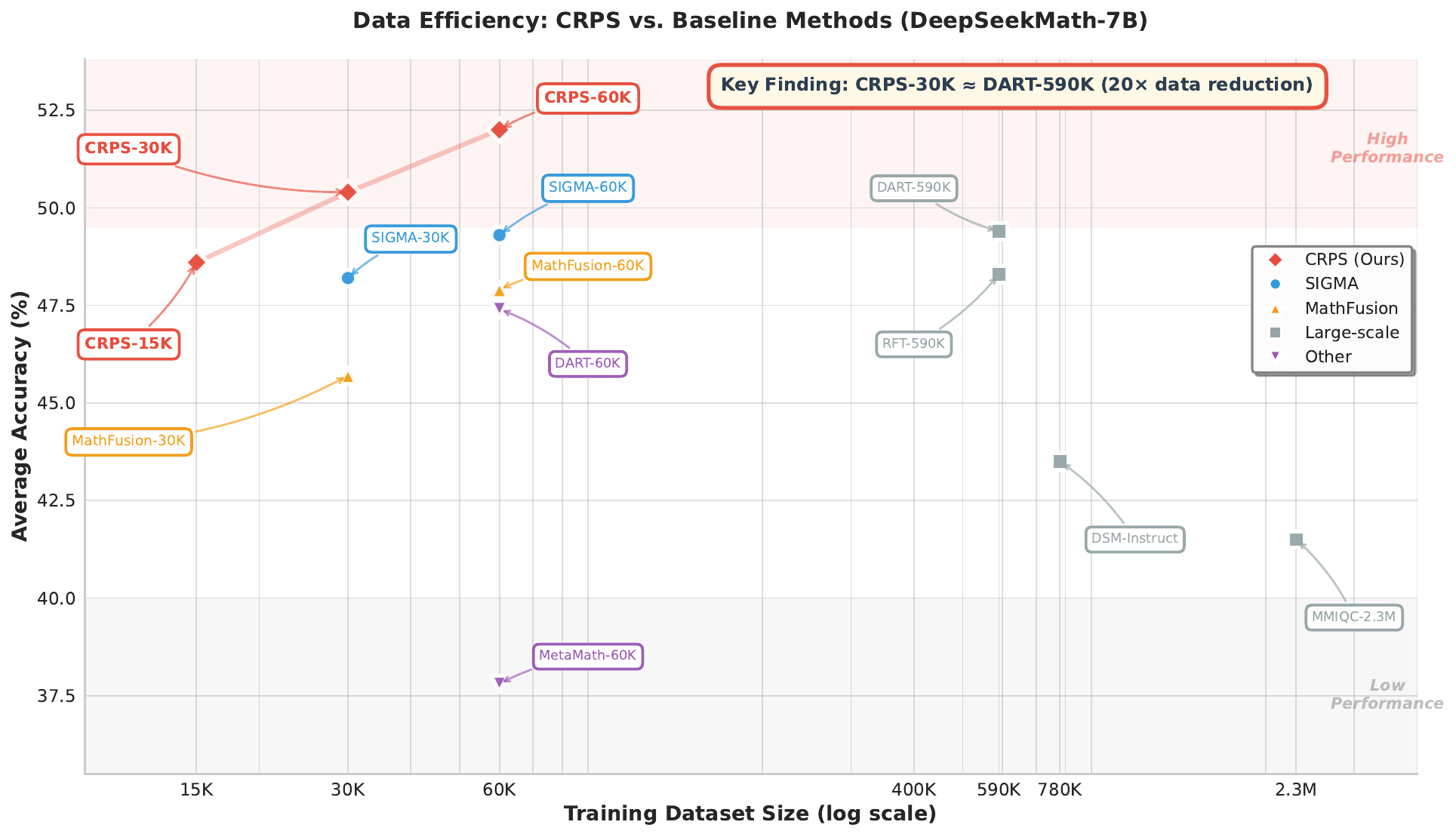}
\caption{An illustration of the reasoning performance of fully fine-tuned DeepSeekMath-7B models. The models are trained with data sizes ranging from 15K to 2.3M.}
\label{fig:overall_performance}
\end{figure*}

\begin{table*}[t]
\centering
\small
\setlength{\tabcolsep}{3pt}
\renewcommand{\arraystretch}{1.0}
\begin{tabular}{@{}l c cc cccc c@{}}
\toprule
\multirow{2}{*}{\textbf{Model}} & \multirow{2}{*}{\textbf{\#Samples}} & \multicolumn{2}{c}{\textbf{In-Domain}} & \multicolumn{4}{c}{\textbf{Out-of-Domain}} & \multirow{2}{*}{\textbf{Avg.}} \\
\cmidrule(lr){3-4} \cmidrule(lr){5-8}
& & MATH & GSM8K & College & DM & Olympiad & Theorem & \\
\midrule
\multicolumn{9}{c}{\textit{LLaMA3-8B (General-Purpose Base Model)}} \\
\midrule
LLaMA3-MetaMath & 400K & 32.5 & 77.3 & 20.6 & 35.0 & 5.5 & 13.8 & 30.8 \\
LLaMA3-RFT & 590K & 39.7 & 81.7 & 23.9 & 41.7 & 9.3 & 14.9 & 35.2 \\
LLaMA3-DART & 590K & 46.6 & 81.1 & 28.8 & 48.0 & 14.5 & 19.4 & 39.7 \\
LLaMA3-MMIQC & 2.3M & 39.5 & 77.6 & 29.5 & 41.0 & 9.6 & 16.2 & 35.6 \\
\midrule
\rowcolor{gray!15}
\textbf{LLaMA3-CRPS-15K} & \textbf{15K} & 37.8$^{*}$ & 83.2$^{*}$ & 25.8$^{*}$ & 44.5$^{*}$ & 11.9$^{*}$ & 23.5$^{*}$ & 37.8$^{*}$ \\
\midrule
MathFusion (Sequential) & 30K & 38.8 & 77.9 & 25.1 & 42.0 & 12.6 & 17.0 & 35.6 \\
\rowcolor{blue!8}
SIGMA-30K & 30K & 40.8 & 79.5 & 26.3 & 47.5 & 12.7 & 19.1 & 37.7 \\
\rowcolor{gray!15}
\textbf{LLaMA3-CRPS-30K} & \textbf{30K} & \textbf{42.1}$^{*}$\textsuperscript{\textcolor{DarkGreen}{+1.3}} & \textbf{81.8}$^{*}$\textsuperscript{\textcolor{DarkGreen}{+2.3}} & \textbf{27.9}$^{*}$\textsuperscript{\textcolor{DarkGreen}{+1.6}} & \textbf{49.2}$^{*}$\textsuperscript{\textcolor{DarkGreen}{+1.7}} & \textbf{14.8}$^{*}$\textsuperscript{\textcolor{DarkGreen}{+2.1}} & \textbf{21.6}$^{*}$\textsuperscript{\textcolor{DarkGreen}{+2.5}} & \textbf{39.6}$^{*}$\textsuperscript{\textcolor{DarkGreen}{+1.9}} \\
\midrule
LLaMA3-DART & 60K & 39.6 & 82.2 & 27.9 & 39.9 & 12.9 & 22.9 & 37.6 \\
MathFusion & 60K & 46.5 & 79.2 & 27.9 & 43.4 & 17.2 & 20.0 & 39.0 \\
\rowcolor{blue!8}
SIGMA-60K & 60K & 44.9 & 82.4 & 28.1 & 49.2 & 15.3 & 21.3 & 40.2 \\
\rowcolor{gray!15}
\textbf{LLaMA3-CRPS-60K} & \textbf{60K} & \textbf{46.8}$^{*}$\textsuperscript{\textcolor{DarkGreen}{+1.9}} & \textbf{83.5}$^{*}$\textsuperscript{\textcolor{DarkGreen}{+1.1}} & \textbf{29.4}$^{*}$\textsuperscript{\textcolor{DarkGreen}{+1.3}} & \textbf{51.3}$^{*}$\textsuperscript{\textcolor{DarkGreen}{+2.1}} & \textbf{17.9}$^{*}$\textsuperscript{\textcolor{DarkGreen}{+2.6}} & \textbf{23.8}$^{*}$\textsuperscript{\textcolor{DarkGreen}{+2.5}} & \textbf{41.9}$^{*}$\textsuperscript{\textcolor{DarkGreen}{+1.7}} \\
\midrule
\multicolumn{9}{c}{\textit{Mistral-7B-v0.1 (General-Purpose Base Model)}} \\
\midrule
Mistral-MetaMath & 400K & 29.8 & 76.5 & 19.3 & 28.0 & 5.9 & 14.0 & 28.9 \\
Mistral-WizardMath & 418K & 32.3 & 80.4 & 23.1 & 38.4 & 7.7 & 16.6 & 33.1 \\
Mistral-RFT & 590K & 38.7 & 82.3 & 24.2 & 35.6 & 8.7 & 16.2 & 34.3 \\
Mistral-DART & 590K & 45.5 & 81.1 & 29.4 & 45.1 & 14.7 & 17.0 & 38.8 \\
\midrule
\rowcolor{gray!15}
\textbf{Mistral-CRPS-15K} & \textbf{15K} & 31.8$^{*}$ & 76.9$^{*}$ & 22.4$^{*}$ & 41.2$^{*}$ & 9.1$^{*}$ & 17.8$^{*}$ & 33.2$^{*}$ \\
\midrule
MathFusion (Sequential) & 30K & 32.7 & 73.9 & 18.9 & 29.3 & 9.3 & 15.5 & 29.9 \\
\rowcolor{blue!8}
SIGMA-30K & 30K & 35.5 & 78.6 & 22.1 & 43.8 & 11.1 & 18.0 & 34.9 \\
\rowcolor{gray!15}
\textbf{Mistral-CRPS-30K} & \textbf{30K} & \textbf{37.2}$^{*}$\textsuperscript{\textcolor{DarkGreen}{+2.3}} & \textbf{80.1}$^{*}$\textsuperscript{\textcolor{DarkGreen}{+1.5}} & \textbf{24.3}$^{*}$\textsuperscript{\textcolor{DarkGreen}{+2.2}} & \textbf{46.8}$^{*}$\textsuperscript{\textcolor{DarkGreen}{+3.0}} & \textbf{12.5}$^{*}$\textsuperscript{\textcolor{DarkGreen}{+1.4}} & \textbf{19.7}$^{*}$\textsuperscript{\textcolor{DarkGreen}{+1.7}} & \textbf{36.8}$^{*}$\textsuperscript{\textcolor{DarkGreen}{+1.9}} \\
\midrule
Mistral-MetaMath & 60K & 22.7 & 70.8 & 14.1 & 27.2 & 5.0 & 12.2 & 25.3 \\
Mistral-DART & 60K & 34.1 & 77.2 & 23.4 & 36.0 & 8.7 & 18.2 & 32.9 \\
MathFusion & 60K & 41.6 & 79.8 & 24.3 & 39.2 & 13.6 & 18.1 & 36.1 \\
\rowcolor{blue!8}
SIGMA-60K & 60K & 40.3 & 79.2 & 24.1 & 46.1 & 12.3 & 19.2 & 36.9 \\
\rowcolor{gray!15}
\textbf{Mistral-CRPS-60K} & \textbf{60K} & \textbf{42.7}$^{*}$\textsuperscript{\textcolor{DarkGreen}{+2.4}} & \textbf{81.4}$^{*}$\textsuperscript{\textcolor{DarkGreen}{+2.2}} & \textbf{26.1}$^{*}$\textsuperscript{\textcolor{DarkGreen}{+2.0}} & \textbf{48.9}$^{*}$\textsuperscript{\textcolor{DarkGreen}{+2.8}} & \textbf{14.2}$^{*}$\textsuperscript{\textcolor{DarkGreen}{+1.9}} & \textbf{21.0}$^{*}$\textsuperscript{\textcolor{DarkGreen}{+1.8}} & \textbf{39.1}$^{*}$\textsuperscript{\textcolor{DarkGreen}{+2.2}} \\
\bottomrule
\end{tabular}
\caption{Performance comparison across base models, training methods, and dataset scales.
Arrows indicate accuracy changes relative to the strongest baseline (highlighted in blue). Best results in each data scale are in \textbf{bold}. $^{*}$ indicate statistical significance at $p < 0.05$ compared to the best baseline (calculated via paired t-test).}
\label{tab:main-results}
\end{table*}

\begin{table}[t]
\centering
\small
\setlength{\tabcolsep}{6pt}
\renewcommand{\arraystretch}{1.1}
\begin{tabular}{@{}l cc@{}}
\toprule
\textbf{Method} & \textbf{HumanEval} & \textbf{StrategyQA} \\
\midrule
\multicolumn{3}{l}{\textit{Code Generation (Base: Qwen2.5-Coder-7B)}} \\
Supervised FT & 48.2 & -- \\
Best-of-32 Sampling & 52.4 & -- \\
Self-Refinement & 54.1 & -- \\
\textbf{CRPS (Ours)} & \textbf{57.9} & -- \\
\midrule
\multicolumn{3}{l}{\textit{Commonsense (Base: LLaMA3-8B)}} \\
Supervised FT & -- & 68.3 \\
Best-of-32 Sampling & -- & 71.5 \\
Self-Refinement & -- & 73.2 \\
\textbf{CRPS (Ours)} & -- & \textbf{75.8} \\
\bottomrule
\end{tabular}
\caption{Performance on non-mathematical reasoning tasks. We report pass@1 accuracy for HumanEval and exact-match accuracy for StrategyQA. CRPS consistently improves performance over standard refinement methods by leveraging the contrast between valid and invalid execution traces.}
\label{tab:non-math}
\end{table}

\section{Implementation Details}
\label{sec:implementation_details}

\subsection{CRPS Data Synthesis Implementation}

In alignment with the decoupled explorer-analyst paradigm described in Section~\ref{sec:method}, our data synthesis pipeline utilizes distinct models for exploration and critique. Specifically, we employ \textbf{Qwen2.5-Math-7B-Instruct} as the explorer to generate the candidate trajectories, and \textbf{gpt-5-mini} as the analyst for generating critiques and synthesizing the final reasoning paths. This ensures the supervision signals are derived from a highly capable teacher model that possesses advanced meta-cognitive abilities.

\paragraph{Exploratory Trajectory Collection (MCTS).}
For each problem $q_i$ in the seed dataset, we employ the Upper Confidence Bound for Trees (UCT) \cite{vodopivec2014enhancing} algorithm using the explorer model. We set the exploration constant $c_{\text{puct}}=1.4$ to encourage diverse reasoning strategies. The maximum tree depth is limited to 16 steps, with a maximum of $k=3$ actions sampled per node. We perform 10 rollouts per problem. Terminal states are rewarded binary values ($r=1$ for correct answers, $r=0$ otherwise) based on strict ground-truth matching.

\paragraph{Trajectory Stratification and Pairing.}
From the constructed trees, we extract complete trajectories and stratify them into:
\begin{itemize}
    \item $\mathcal{T}^{\text{high}}$: Paths yielding the correct answer with the highest accumulated process reward (or shortest length among correct paths).
    \item $\mathcal{T}^{\text{low}}$: Paths yielding incorrect answers ($\tau_{incorrect}$), or correct but inefficient/suboptimal paths ($\tau_{inefficient}$) as described in Section~\ref{sec:mtcs_construct}.
\end{itemize}
We sample $K=10$ contrastive pairs $(\tau^+, \tau^-)$ per problem to serve as inputs for the analyst model. Crucially, consistent with Section~\ref{sec:mtcs_construct}, $\tau^-$ is sampled proportionally to its visit count $N(\tau)$ to target the explorer's stubborn error modes, rather than uniformly at random.

\paragraph{Synthesis and Dataset Construction.}
Following the synthesis of candidate reasoning paths by the analyst (Section~\ref{sec:synthesis}), we apply the verification function $\mathcal{V}(\hat{\tau})$. For mathematical tasks, this involves extracting the final answer and performing an exact-match comparison with the ground truth (using SymPy \cite{meurer2017sympy} for equivalence checks). Only synthesized paths that yield the correct answer are retained for the training pool. To construct the final datasets $\mathcal{D}_{\text{syn}}$ at different scales (15K, 30K, 60K) for our scaling laws analysis, we perform uniform sampling from the verified pool. 
For our primary \textbf{CRPS-30K} dataset, this corresponds to retaining approximately 2 synthesized variations per seed problem from the combined GSM8K and MATH training sets.

The specific prompts for critique and synthesis are detailed in Appendix~\ref{sec:prompts}.

\subsection{Baselines}
\label{sec:baselines}
    \textbf{Vanilla MCTS / Rejection Sampling (RFT)}~\cite{yuan2023scaling}: We construct the dataset by selecting the single highest-reward trajectory found during the explorer's MCTS for each problem. This represents the standard ``selection-based'' paradigm.
    
    \textbf{MMIQC}~\cite{liu2025augmenting}: A massive multi-source instruction tuning dataset containing over 2.3M examples. We include this to compare CRPS against massive-scale data scaling.
    
    \textbf{DART-Math}~\cite{tong2024dartmath}: A strong baseline that applies difficulty-aware rejection sampling. We re-implement their difficulty scoring mechanism to select the top 590K examples from our search trees. Note that while we utilize the same raw trajectory pool for experimental consistency, DART-Math requires retaining this massive volume to achieve its performance, implying a significantly higher exploration budget compared to the sparse subset required by CRPS.
    
    \textbf{SIGMA}~\cite{ren2025sigma}: A recent feedback-driven refinement method. We implement a fair offline adaptation of SIGMA. Using the same MCTS trajectories as CRPS, we construct training pairs $(q, \tau^-) \to \tau^+$ where the target model learns to rewrite a suboptimal path into a correct one, matching the dataset size of our CRPS experiments (30K/60K). This isolates the impact of the contrastive synthesis objective versus direct refinement.

    \textbf{MathFusion}~\cite{pei2025mathfusion}: We compare against the sequential training setup of MathFusion, which aggregates diverse reasoning components.

\subsection{Models and Training}
\label{sec:basemodels}
We conduct experiments on three representative target base models to demonstrate architectural robustness and transferability:
\begin{itemize}
    \item \textbf{DeepSeekMath-7B-Base}~\cite{shao2024deepseekmath}: Specialized in mathematical reasoning.
    \item \textbf{LLaMA3-8B}~\cite{grattafiori2024llama}: A strong general-purpose foundation model.
    \item \textbf{Mistral-7B-v0.1}~\cite{jiang2023mistral7b}: Widely used for its sliding window attention efficiency.
\end{itemize}

All target models are full fine-tuned using the standard causal language modeling objective. 
The training data consists of the problem statement $q$ as the prompt and the synthesized reasoning path $\hat{\tau}$ as the completion. We use the AdamW optimizer with $\beta_1=0.9, \beta_2=0.95$. The learning rate is warmed up for 3\% of the total steps to a peak of $2 \times 10^{-5}$ and then decays to zero via a cosine schedule. We use a global batch size of 128. Training is performed on 8 $\times$ NVIDIA A800 (80GB) GPUs using DeepSpeed ZeRO-3 \cite{rasley2020deepspeed} offloading. Models are trained for 3 epochs.

\subsection{Evaluation Protocol}

To ensure reproducibility, we employ a standardized evaluation protocol across all experiments.
\begin{itemize}
    \item \textbf{Inference:} We use zero-shot Chain-of-Thought (CoT) prompting. To eliminate randomness and measure the model's robust capability, we use greedy decoding ($\text{temperature}=0$) for all main results.
    \item \textbf{Answer Extraction:} We extract the final answer from the model's output (typically boxed or last numerical value) and perform exact match comparison against the ground truth. For mathematical equivalence (e.g., fractions vs. decimals), we utilize the SymPy library for symbolic verification.
\end{itemize}

\section{Qualitative Analysis}
\label{sec:qualitative_appendix}

To understand the mechanisms driving the quantitative gains, we perform a qualitative dissection of the synthesized reasoning paths.

\subsection{Mechanisms of Contrastive Correction}
We analyze representative cases (detailed in Appendix~\ref{sec:examples}) to understand how contrastive signals drive reasoning improvements. The synthesis process demonstrates a consistent shift from fragile heuristics to structural rigor, driven by the explicit verbalization of failure modes.

\begin{itemize}
    \item \textbf{From Ad-hoc Formulas to First Principles (Case 1, Combinatorics):} In the triangular peg board problem, the baseline MCTS fell into a ``Grid Assumption'' trap, treating the triangular board as a rectangular grid and computing factorials for each color. The contrastive critique identified this as a geometric mismatch: positions only exist where $r+c \leq 6$. Consequently, the synthesized path explicitly verbalized the strategic pivot: \textit{``It is tempting to multiply factorials as if the board were rectangular, but on this triangular board, the constraints at each step leave exactly one valid column.''}  This demonstrates the model's ability to inhibit high-probability heuristic tokens in favor of constraint-aware enumeration.

    \item \textbf{From Brute-Force Calculus to Structural Insight (Case 2, Algebra):} We observe that CRPS also optimizes reasoning efficiency (Soft Negatives). In the quadratic minimization problem, while the baseline correctly attempted the task by setting partial derivatives to zero and solving a parametric linear system, the contrastive analysis flagged this as computationally heavy and prone to stalling. The synthesized CRPS solution bypassed this approach by completing the square directly, thereby revealing the minimum-value condition structurally and reducing algebraic complexity.

    \item \textbf{From Unjustified Shortcuts to Rigorous Integration (Case 3, Calculus):} For the cylindrical wedge volume problem, the baseline assumed the wedge was simply half a cylinder---arriving at the correct answer through fundamentally flawed reasoning. The contrastive analyst identified this as an unjustified geometric simplification. The resulting synthesized path set up coordinates at the tangent point and performed proper integration, explicitly stating: \textit{``A common mistake is to assume the wedge is simply half a cylinder---but each slice has a different height, so integration is necessary.''}
\end{itemize}

These transformations confirm that CRPS does not merely filter for correctness; it utilizes the distribution of suboptimal paths to condition generation against specific logical leaps and heuristic traps, resulting in reasoning chains that are both correct and explicitly justified.

\section{Generalization to Non-Mathematical Reasoning}
\label{sec:non_math_generalization}

To validate the universality of the CRPS framework, we extend our experiments to code generation (\textbf{HumanEval}~\cite{chen2021evaluating}) and multi-hop commonsense reasoning (\textbf{StrategyQA}~\cite{geva2021strategyqa}). We adapt the MCTS reward mechanism $r(\tau)$ to domain-specific verifiers: execution results against unit tests for code, and exact answer matching for commonsense queries. We adapt the decoupled architecture to these domains by employing \textbf{Qwen2.5-Coder-7B}~\cite{hui2024qwen2} and \textbf{LLaMA3-8B} as the domain-specific explorers (and subsequent target base models), while retaining \textbf{gpt-5-mini} as the analyst for contrastive critique and synthesis. We compare CRPS against Supervised Fine-Tuning (SFT), Best-of-32 Sampling, and iterative Self-Refinement~\cite{madaan2023selfrefine}.

As shown in Table~\ref{tab:non-math}, CRPS consistently outperforms baselines across both domains. On HumanEval, CRPS achieves 57.9\% pass@1, surpassing Self-Refinement by 3.8 points. Qualitative analysis suggests that the contrastive mechanism effectively targets latent failure modes: in code generation, the explorer's $\mathcal{T}^{\text{low}}$ trajectories often expose edge-case bugs, while in StrategyQA, they reveal ``reasoning shortcuts'' where correct answers are derived from hallucinated logic. By enabling the analyst to synthesize paths conditioned on explicitly avoiding these structural pitfalls, CRPS demonstrates that the principle of learning from distilled contrastive search trajectories transfers effectively beyond mathematical formalism.

\begin{table}[t]
\centering
\small
\setlength{\tabcolsep}{2pt}
\renewcommand{\arraystretch}{1.1}
\begin{tabular}{@{}l c c c@{}}
\toprule
\multirow{2}{*}{\textbf{Method}} & \textbf{Target} & \textbf{Dataset}  & \textbf{Training}  \\
& \textbf{Acc.} & \textbf{Size} & \textbf{Cost (h)}  \\
\midrule
\textit{Baselines} \\
DART-Math & 49.4\% & 590K  & 192  \\
RFT & 48.3\% & 590K  & 192  \\
SIGMA & 48.2\% & 30K  & 10  \\
\midrule
\textit{Ours} \\
\rowcolor{gray!10}
\textbf{CRPS} & \textbf{50.4\%} & \textbf{30K}  & \textbf{10}  \\
\bottomrule
\end{tabular}
\caption{Computational cost breakdown to achieve $\approx$50\% accuracy on MATH+GSM8K (DeepSeekMath-7B).
CRPS shifts the compute burden from massive exploration and training to targeted synthesis, resulting in a \textbf{19.2$\times$ reduction} in training GPU hours compared to standard search-based selection.}
\label{tab:compute_efficiency}
\end{table}

\section{Step Segmentation Protocol}
\label{sec:segmentation_protocol}

As discussed in Section~\ref{sec:method}, defining the atomic unit of reasoning is a fundamental challenge in CoT research. We define a ``step'' as a \textit{Semantic Reasoning Act}---a discrete move that transforms the state of the problem. To ensure consistent branching for MCTS and semantic alignment for CRPS, we employ a hierarchical segmentation strategy, prioritizing structural delimiters over linguistic ones (Table~\ref{tab:segmentation_rules}). We strictly avoid splitting within mathematical expressions (e.g., inside \LaTeX{} \texttt{\$...\$}) to preserve equation integrity, and enforce a maximum length of 256 tokens per step to prevent ``run-on'' steps that dilute the credit assignment signal.

\begin{table*}[t]
\centering
\small
\setlength{\tabcolsep}{4pt}
\renewcommand{\arraystretch}{1.2}
\begin{tabular}{@{}c l p{3.5cm} p{5.5cm}@{}}
\toprule
\textbf{Priority} & \textbf{Delimiter Type} & \textbf{Specific Tokens / Patterns} & \textbf{Rationale} \\
\midrule
\textbf{1} & \textbf{Structure} (Primary) & \texttt{\textbackslash n}, \texttt{\textbackslash n\textbackslash n}, \texttt{\textbackslash\textbackslash} (\LaTeX{} newline) & Natural boundaries in math derivations; ensures semantic completeness. \\
\textbf{2} & \textbf{Logic Connectors} & \texttt{Therefore,}, \texttt{Thus,}, \texttt{Hence,}, \texttt{So,}, \texttt{Consequently,} & Signals a deductive leap or conclusion; critical for value estimation ($Q(s)$). \\
\textbf{3} & \textbf{Explicit Enumeration} & \texttt{1.}, \texttt{2.}, \texttt{Step 1:}, \texttt{First,} & Explicitly structured reasoning steps in CoT. \\
\bottomrule
\end{tabular}
\caption{Heuristic rules for step segmentation in MCTS and CRPS semantic alignment.}
\label{tab:segmentation_rules}
\end{table*}

\section{Semantic Stability Evaluation Setup}
\label{sec:stability_setup}

To rigorously evaluate the semantic stability of the fine-tuned target models (Section~\ref{sec:generalization}), we constructed a perturbed dataset based on the GSM8K test set.

\subsection{Data Construction}
We randomly sampled 200 problems from the GSM8K test set as seed queries. For each seed problem $q$, we utilized GPT-5 to generate a perturbed variant $q'$ using the following prompt template. The generation was constrained to introduce two types of semantic noise:
\begin{enumerate}
    \item \textbf{Linguistic Paraphrasing:} Altering sentence structures and replacing entities (e.g., names, objects) while preserving the numerical relationships.
    \item \textbf{Distractor Injection:} Inserting irrelevant context sentences that do not affect the calculation but serve as ``attention traps''.
\end{enumerate}

\begin{tcolorbox}[colback=gray!10!white, colframe=gray!50!black, title=Perturbation Generation Prompt]
\textbf{Instruction:} 
Rewrite the following math word problem to create a ``Semantically Equivalent but Perturbed'' version. 
1. Change the names of people and objects.
2. Rephrase the sentence structures significantly.
3. Insert one sentence of ``distractor'' information that contains a number but is irrelevant to the solution.
4. Do NOT change the underlying logic or the required calculation flow. The final answer must remain exactly the same.

\textbf{Input Problem:} 
[Insert Original Problem]

\textbf{Output Problem:}
\end{tcolorbox}

\subsection{Metric Definition}
We employ \textbf{Strict Consistency} as our primary metric. Let $M(q)$ denote the binary correctness (1 for correct, 0 for incorrect) of the target model on query $q$. For a dataset of size $N$, the consistency score $S$ is calculated as:
\begin{equation}
    S = \frac{1}{N} \sum_{i=1}^{N} \mathbb{I}[M(q_i) = 1 \land M(q'_i) = 1]
\end{equation}
This metric is stricter than simple answer agreement (where two wrong answers might count as consistent) or average accuracy. It specifically measures the target model's ability to maintain correct reasoning despite adversarial surface-level perturbations, verifying whether it has truly internalized the robust structural logic distilled during the CRPS synthesis phase.

\section{Prompts and Instructions}
\label{sec:prompts}

To ensure reproducibility, we provide the exact prompt templates used in our CRPS framework. The process strictly follows the dual-granularity analysis described in Section~\ref{sec:contrastive}, separating global strategic insights from local step-wise critiques performed by the analyst model.

\subsection{Phase 1: Dual-Granularity Contrastive Analysis}
\label{sec:prompt_critique}

In this phase, the analyst model (\textbf{gpt-5-mini}) processes the outputs from the explorer. It receives the problem $q$, a high-quality trajectory $\tau^+$ (positive), and a low-quality trajectory $\tau^-$ (negative) generated during MCTS. The instructions explicitly require the analyst to perform \textbf{Semantic Alignment} to locate the divergence step and generate both global and local critiques.

\begin{tcolorbox}[colback=gray!10!white, colframe=gray!50!black, breakable, title=System Prompt: The Contrastive Analyst]
You are an expert Mathematical Reasoning Analyst. Your task is to perform a dual-granularity contrastive analysis between two reasoning trajectories for the same mathematical problem. One trajectory arrives at the correct answer (Trajectory A) and the other arrives at an incorrect answer (Trajectory B). You must identify where and why the trajectories diverge, analyze both local step-level logic and global strategic differences, and synthesize actionable guidance.
\end{tcolorbox}

\begin{tcolorbox}[colback=white, colframe=gray!50!black, breakable, title=User Prompt: Dual-Granularity Analysis Instruction]
\textbf{Problem:}
[INSERT PROBLEM $q$]

\textbf{Trajectory A (Correct):}
[INSERT TRAJECTORY $\tau^+$]

\textbf{Trajectory B (Incorrect):}
[INSERT TRAJECTORY $\tau^-$]

Perform a dual-granularity contrastive analysis of the two trajectories above. Provide your analysis as a JSON object with the following structure:

\begin{lstlisting}[language=json, basicstyle=\small\ttfamily, breaklines=true]
{
  "divergence_step_index": <int>,
  "local_step_critique": {
      "trajectory_a_logic": "<description of Trajectory A's
          reasoning at the divergence point>",
      "trajectory_b_logic": "<description of Trajectory B's
          reasoning at the divergence point>",
      "critique_of_difference": "<precise explanation of why
          Trajectory B's step is incorrect and how Trajectory
          A's step is correct>"
  },
  "global_strategic_analysis": "<high-level comparison of
      the overall strategies employed by each trajectory>",
  "synthesized_guidance": {
      "success_pattern": "<the key reasoning pattern from
          Trajectory A that led to the correct answer>",
      "failure_mode_to_avoid": "<the specific reasoning
          pitfall from Trajectory B to be avoided>"
  }
}
\end{lstlisting}

Return ONLY the JSON object, with no additional text.
\end{tcolorbox}

\subsection{Phase 2: Pattern-Informed Path Synthesis}
\label{sec:prompt_synthesis}

In this phase, the analyst model synthesizes a new reasoning path $\hat{\tau}$. The generation is conditioned on the extracted critiques, acting as a prompt-based regularizer to explicitly navigate around the explorer's identified failure mode.

\begin{tcolorbox}[colback=gray!10!white, colframe=gray!50!black, breakable, title=System Prompt: The Pattern-Informed Solver]
You are an advanced Mathematical Reasoning Engine. Your task is to solve a mathematical problem by generating a step-by-step solution that is informed by prior contrastive analysis of correct and incorrect reasoning paths.

\textbf{KEY REQUIREMENTS:}
\\
1. \textbf{INCORPORATE} contrastive insights naturally into your reasoning. At critical steps, briefly explain WHY you chose the correct approach and what common mistake to avoid. For example: ``A common mistake is to use $C(6,2)$ here, which treats identical objects as distinct. Instead, we enumerate cases.'' This is the core value of CRPS.
\\
2. Do \textbf{NOT} use meta-language like ``the critique suggests'', ``following the success pattern'', or ``as identified in the analysis''. Write as if \textbf{YOU} discovered these insights while solving.
\\
3. Match the target model's formatting style (bold headers, \LaTeX, numbered steps).
\end{tcolorbox}

\begin{tcolorbox}[colback=white, colframe=gray!50!black, breakable, title=User Prompt: Synthesis Instruction]
\textbf{Problem:}
[INSERT PROBLEM $q$]

\textbf{Contrastive Insights} (weave naturally into solution, do NOT reference directly):
\begin{itemize}
    \item Why the correct approach works: [INSERT \texttt{success\_pattern}]
    \item Common mistake to avoid: [INSERT \texttt{failure\_mode\_to\_avoid}]
    \item Key difference at Step [INSERT \texttt{divergence\_step\_index}]: [INSERT \texttt{critique\_of\_difference}]
    \item Strategic overview: [INSERT \texttt{global\_strategic\_analysis}]
\end{itemize}

\textbf{Target Output Style:}
[INSERT style example from a successful trajectory]

Solve step by step. At the critical decision point, naturally explain why you chose your approach and what pitfall you are avoiding — as if you discovered this yourself. Do NOT say ``the critique'' or ``the analysis''.

\textbf{Format:}
Step 1: ...
Step 2: ...
...
Final Answer: \textbackslash boxed\{...\}
\end{tcolorbox}

\subsection{Output Parsing and Robustness}
\label{sec:parsing_robustness}

To ensure the autonomy of our pipeline regardless of the chosen analyst, we implement a lightweight post-processing mechanism. We utilize regular expressions to extract the JSON block from the model's raw output and apply a rule-based parser to correct common formatting issues (e.g., unescaped quotes or missing trailing braces). Instances that remain unparsable after this repair process are discarded. Empirically, we observe a parsing failure rate of less than 5\% across all evaluated models, confirming that the instruction-tuned models possess sufficient structural adherence to support our automated synthesis loop without human intervention.

\section{Examples}
\label{sec:examples}

Here we demonstrate the CRPS pipeline on three representative cases. To showcase the framework's versatility, we include both \textbf{Correctness Contrasts} (learning from errors, Cases 1 \& 3) and an \textbf{Efficiency Contrast} (learning from suboptimal strategies, Case 2). The input trajectories ($\tau^+$ and $\tau^-$) are sampled from the exploration trees generated by the explorer model (\textbf{Qwen2.5-Math-7B-Instruct}). The synthesized outputs reflect the analyst model's (\textbf{gpt-5-mini}) actual generation capabilities, showing how it incorporates its own contrastive critiques into procedural steps without excessive meta-cognitive language. These distilled, high-quality paths are subsequently used to train the target base models.

Case~1 illustrates a \emph{wrong mental model}: the explorer confidently applies a rectangular-grid counting strategy to a triangular board, and the contrastive analysis pinpoints the geometric mismatch at the very first step. Case~2 demonstrates an \emph{efficiency gap}: both trajectories reach the correct answer, but the positive path uses an elegant structural approach (completing the square) while the negative path stalls in heavy parametric algebra. Case~3 exposes a \emph{geometric reasoning error}: the explorer incorrectly assumes a wedge is half a cylinder, and the critique identifies the need for integration over cross-sectional slices of varying height.

\begin{tcolorbox}[colback=red!5!white, colframe=red!75!black, breakable, title=Case 1. Hard Negative: Wrong Mental Model (Combinatorics)]
\textbf{Question:} There are 5 yellow pegs, 4 red pegs, 3 green pegs, 2 blue pegs, and 1 orange peg to be placed on a triangular peg board. In how many ways can the pegs be placed so that no (horizontal) row or (vertical) column contains two or more pegs of the same color?

\vspace{0.5em}
\textbf{Input Trajectories:}
\begin{itemize}
    \item $\tau^+$ (Positive): Maps the triangular board structure, identifies that position $(r,c)$ exists iff $r+c \leq 6$, and shows each color placement is uniquely forced.
    \item $\tau^-$ (Hard Negative): Treats the triangular board as a rectangular grid and computes $5! \times 4! \times 3! \times 2! \times 1!$, vastly overcounting.
\end{itemize}

\vspace{0.5em}
\textbf{Phase 1: Dual-Granularity Critique (JSON)}
\begin{lstlisting}[language=json, basicstyle=\small\ttfamily, breaklines=true]
{
  "divergence_step_index": 1,
  "local_step_critique": {
      "trajectory_a_logic": "Maps out exact positions in each row and column of the triangular board before counting.",
      "trajectory_b_logic": "Treats the board as a full rectangular grid, computing factorials for each color.",
      "critique_of_difference": "The triangular board has position (r,c) only if r+c<=6. Trajectory B assumes all n! permutations are valid, but most positions do not exist on the triangular board."
  },
  "global_strategic_analysis": "Both attempt sequential color-by-color placement. Trajectory B ignores the geometric constraints, while Trajectory A recognizes that these constraints uniquely force each placement.",
  "synthesized_guidance": {
      "success_pattern": "Map out exact positions available in each row and column before counting. Recognize when constraints are tight enough to force a unique arrangement.",
      "failure_mode_to_avoid": "Treating a triangular board as a rectangular grid when computing permutations."
  }
}
\end{lstlisting}

\vspace{0.5em}
\textbf{Phase 2: Pattern-Informed Synthesis}
\\
\textbf{Step 1:} Understand the board structure.
The triangular board has position $(r, c)$ iff $r + c \leq 6$. Row $r$ has $6 - r$ positions; column $c$ has $6 - c$ positions. Total: 15 positions for 15 pegs.
\\
\textbf{Step 2:} Place yellow pegs (5 pegs).
Each of 5 rows and 5 columns needs exactly one yellow peg. We need a bijection $\sigma:\{1,\ldots,5\}\to\{1,\ldots,5\}$ with $\sigma(r) \leq 6-r$. Working upward: $\sigma(5)=1$ (forced), $\sigma(4)=2$ (forced), $\sigma(3)=3$, $\sigma(2)=4$, $\sigma(1)=5$. Exactly 1 way.
\\
\textbf{Step 3:} Place remaining colors.
By the same forced-assignment argument applied to red ($\tau$), green ($\gamma$), blue ($\beta$), and orange: each is uniquely determined. It is tempting to multiply factorials as if the board were rectangular, but on this triangular board, the constraints at each step leave exactly one valid column.
\\
\textbf{Step 4:} Count.
Total arrangements = $1 \times 1 \times 1 \times 1 \times 1 = 1$.
\\
\textbf{Final Answer:} $\boxed{1}$
\end{tcolorbox}

\begin{tcolorbox}[colback=blue!5!white, colframe=blue!75!black, breakable, title=Case 2. Soft Negative: Efficiency Gap (Algebra)]
\textbf{Question:} There exists a constant $k$ so that the minimum value of
$4x^2 - 6kxy + (3k^2 + 2)y^2 - 4x - 4y + 6$
over all real numbers $x$ and $y$ is 0. Find $k$.

\vspace{0.5em}
\textbf{Input Trajectories:}
\begin{itemize}
    \item $\tau^+$ (Positive): Completes the square to express the function as a sum of squared terms plus a constant, directly revealing conditions on $k$.
    \item $\tau^-$ (Soft Negative): Sets partial derivatives to zero, attempts to solve a parametric $2\times 2$ linear system, but stalls during intermediate algebra.
\end{itemize}

\vspace{0.5em}
\textbf{Phase 1: Dual-Granularity Critique (JSON)}
\begin{lstlisting}[language=json, basicstyle=\small\ttfamily, breaklines=true]
{
  "divergence_step_index": 1,
  "local_step_critique": {
      "trajectory_a_logic": "Analyzes the quadratic form structure and completes the square.",
      "trajectory_b_logic": "Computes partial derivatives and sets up a parametric linear system.",
      "critique_of_difference": "Trajectory B's calculus approach requires inverting a parametric matrix and extensive back-substitution. Trajectory A's structural approach directly reveals when the minimum equals zero."
  },
  "global_strategic_analysis": "Trajectory A adopts a structural approach via completing the square, naturally connecting coefficients to the minimum-value condition. Trajectory B uses brute-force calculus that stalls during algebra.",
  "synthesized_guidance": {
      "success_pattern": "For multivariable quadratic minimization with a parameter, complete the square to express as sum-of-squares plus constant.",
      "failure_mode_to_avoid": "Diving into heavy parametric linear systems without considering structural alternatives."
  }
}
\end{lstlisting}

\vspace{0.5em}
\textbf{Phase 2: Pattern-Informed Synthesis}
\\
\textbf{Step 1:} Choose the method.
Rather than setting partial derivatives to zero and solving a parametric system (which is computationally heavy and prone to stalling), we complete the square directly.
\\
\textbf{Step 2:} Complete the square in $x$.
\[
f(x,y) = 4\!\left(x - \tfrac{6ky+4}{8}\right)^{\!2} + \tfrac{3k^2+8}{4}\,y^2 
\]
\[
- (3k+4)y + 5.
\]
\\
\textbf{Step 3:} Complete the square in $y$.
Since $\frac{3k^2+8}{4} > 0$ for all real $k$, the form is positive definite and a minimum exists. The minimum value is:
\[f_{\min} = 5 - \frac{(3k+4)^2}{3k^2+8}.\]
\\
\textbf{Step 4:} Solve for $k$.
Setting $f_{\min} = 0$:
$5(3k^2 + 8) = (3k + 4)^2 \;\Longrightarrow\; 6k^2 - 24k + 24 = 0 \;\Longrightarrow\; (k - 2)^2 = 0.$
\\
\textbf{Final Answer:} $\boxed{2}$
\end{tcolorbox}

\begin{tcolorbox}[colback=red!5!white, colframe=red!75!black, breakable, title=Case 3. Hard Negative: Geometric Reasoning Error (Calculus)]
\textbf{Question:} A cylindrical log has diameter 12 inches. A wedge is cut from the log by making two planar cuts that go entirely through the log. The first is perpendicular to the axis of the cylinder, and the plane of the second cut forms a $45^\circ$ angle with the plane of the first cut. The intersection of these two planes has exactly one point in common with the log. The number of cubic inches in the wedge can be expressed as $n\pi$. Find $n$.

\vspace{0.5em}
\textbf{Input Trajectories:}
\begin{itemize}
    \item $\tau^+$ (Positive): Sets up coordinates at the tangent point, expresses height as $h(x)=x$, computes chord widths, and integrates.
    \item $\tau^-$ (Hard Negative): Assumes the wedge is half a cylinder, arriving at the correct numerical answer through flawed reasoning.
\end{itemize}

\vspace{0.5em}
\textbf{Phase 1: Dual-Granularity Critique (JSON)}
\begin{lstlisting}[language=json, basicstyle=\small\ttfamily, breaklines=true]
{
  "divergence_step_index": 1,
  "local_step_critique": {
      "trajectory_a_logic": "Sets up a coordinate system at the tangent point and computes cross-sectional slices.",
      "trajectory_b_logic": "Assumes the wedge is half a cylinder without justification.",
      "critique_of_difference": "The 45-degree angled cut tangent to one edge creates a non-trivial shape where each slice has a different height. Trajectory B's simplification is unjustified."
  },
  "global_strategic_analysis": "Trajectory A uses geometric decomposition and integration. Trajectory B takes an unjustified shortcut that accidentally gives the right answer but reflects a fundamental misunderstanding.",
  "synthesized_guidance": {
      "success_pattern": "Set up coordinates at the tangent point, express height as h(x)=x, compute chord widths from the circle equation, and integrate.",
      "failure_mode_to_avoid": "Assuming the wedge is a simple fraction of a cylinder without rigorous justification."
  }
}
\end{lstlisting}

\vspace{0.5em}
\textbf{Phase 2: Pattern-Informed Synthesis}
\\
\textbf{Step 1:} Set up coordinates.
Let the tangent point be the origin, with $x$ running along a diameter from $0$ to $12$ (radius $r=6$, center at $x=6$). The height at position $x$ is $h(x) = x\tan(45^\circ) = x$. The chord width is $w(x) = 2\sqrt{36-(x-6)^2} = 2\sqrt{12x-x^2}$.
\\
\textbf{Step 2:} Set up the integral.
A common mistake is to assume the wedge is simply half a cylinder---but each slice has a different height, so integration is necessary:
\[V = 2\int_0^{12} x\sqrt{12x - x^2}\,dx.\]
\\
\textbf{Step 3:} Evaluate.
Substituting $x = 6 + 6\sin\theta$, $dx = 6\cos\theta\,d\theta$, $\sqrt{12x-x^2} = 6\cos\theta$:
\[V = 432\int_{-\pi/2}^{\pi/2}(1+\sin\theta)\cos^2\theta\,d\theta = \]
\[432\!\left[\frac{\pi}{2} + 0\right] = 216\pi,\]
where the $\sin\theta\cos^2\theta$ term vanishes by odd symmetry. Thus $n = 216$.
\\
\textbf{Final Answer:} $\boxed{216}$
\end{tcolorbox}

\end{document}